%% file: main_paper.tex
\documentclass{article} 
\usepackage{main_paper,times}

\input{math_commands.tex}

\usepackage{hyperref}
\usepackage{url}
\usepackage{algorithm} 
\usepackage{algpseudocode} 
\usepackage{amsmath}
\usepackage{dsfont}
\usepackage{graphicx}

\algdef{SE}[SUBALG]{Indent}{EndIndent}{}{\algorithmicend\ }%
\algtext*{Indent}
\algtext*{EndIndent}

\title{Improving Experience Replay with \\ Successor Representation}

\begin{document}

\maketitle

\begin{abstract}
Prioritized experience replay is a reinforcement learning technique whereby agents speed up learning by replaying useful past experiences. This usefulness is quantified as the expected gain from replaying the experience, a quantity often approximated as the prediction error (TD-error). However, recent work in neuroscience suggests that, in biological organisms, replay is prioritized not only by gain, but also by ``need'' --  a quantity measuring the expected relevance of each experience with respect to the current situation. Importantly, this term is not currently considered in algorithms such as prioritized experience replay. In this paper we present a new approach for prioritizing experiences for replay that considers both gain and need. Our proposed algorithms show a significant increase in performance in benchmarks including the Dyna-Q maze and a selection of Atari games. 
\end{abstract}

\section{Introduction}

Reinforcement learning has achieved impressive results in complex tasks such as Atari games, sometimes outperforming humans \citep{DBLP:journals/nature/MnihKSRVBGRFOPB15}. An important method often used in these results is Q-learning, a ubiquitous algorithm in reinforcement learning (RL). In Q-learning, the agent learns action values (Q-values) on the basis of the reward immediately received and the estimated value of the state resulting from each action. By repeatedly executing each action in each state multiple times, in a trial-and-error fashion, the Q-values can gradually improve over time \citep{sutton_barto_2014}. However, this Q-value estimation process often suffers from severe sample inefficiency, as it often takes an excessive amount of time for the values to converge.

Experience replay is a technique often used in conjunction with Q-learning to improve sample efficiency. Each time the agent takes an action in a state, and observes a reward and a next state, it stores this information (referred to as an experience or an transition) in an array called ``replay buffer''. This allows the agent to sample experiences from the buffer, which, when replayed, lead to Q-values updates. Indeed, Q-learning can be augmented with experience replay, as in the classic tabular Dyna-Q algorithm \citep{sutton_barto_2014}. Similarly, the deep-learning counterpart of Q-learning, DQN, can be greatly improved by using experience replay \citep{DBLP:journals/nature/MnihKSRVBGRFOPB15}. In both cases, experience replay leads to significant improvements in sample efficiency over Q-learning without experience replay. Note, however, that such improvements were first achieved using a simple uniform sampling scheme, whereby each experiences have the same probability of being replayed.

While these algorithms demonstrate that experience replay can accelerate the convergence of the learned values, they also lead to the obvious question: how to prioritize the order of replayed experiences so that the best policy is achieved with the minimum expended computation? While this question has been studied in both tabular and non-tabular (e.g., deep neural networks) domains, there has been no consensus over the best way to prioritize the order replay. Nonetheless, various prioritization heuristics have been proposed. The most common technique replays preferentially experiences associated with a large TD-error, over those associated with a small TD-error. Subsequently, other variants based on different heuristics have also been proposed (e.g., \cite{brittain_bertram_yang_wei_2020}).

In this paper, we propose a novel prioritization scheme for experience replay based on normative principles. The prioritization scheme uses TD-error, as in previous work \citep{schaul_quan_antonoglou_silver_2016}, but augments it with a new metric called the ``need term'', an additional bias term derived normatively and inspired by patterns of replay in biological organisms \citep{mattar_daw_2018}. The need term is obtained from the Successor Representation (SR) \citep{dayan_1993}, characterizing the expected number of future visits to each other state,and thus yielding a natural measure of relevance for each experience in the buffer. We test this new prioritization scheme in different tasks, in combination with both tabular and non-tabular methods, and using baselines from both tabular Dyna-Q \citep{sutton_barto_2014} and DQN \citep{DBLP:journals/nature/MnihKSRVBGRFOPB15}. Our results suggest that our new prioritization scheme significantly improves the existing algorithms.

\section{Related Work}
\label{related work}

Our proposed prioritization method is motivated by prior work in both machine learning and neuroscience. First, from the machine learning side, we rely on the classical prioritization based on prediction error (TD-error), previously used in multiple algorithms including prioritized sweeping (PS) \citep{sutton_barto_2014} and prioritized experience replay (PER) \citep{schaul_quan_antonoglou_silver_2016}. In TD-error prioritization, the probability of sampling an experience is higher if the prediction error expected from that replay -- i.e., the magnitude of the Q-update -- is high. For example, given the experience tuple $t = (s_j, a_j, r_j, s_{j+1}, a_{j+1}, p_j)$, the TD-error is $r_j + \gamma Q(s_{j+1}, a_{j+1}) - Q(a_j, r_j)$, which indicates the extent to which we can learn (change) the Q-value of the pair $(a_j, r_j)$ by replaying this experience. Thus, one can expect that prioritizing experiences based on the TD-error can lead to faster learning and better sample efficiency. Several variants of this prediction error scheme have been proposed, including prioritized sequence experience replay (PSER) \citep{brittain_bertram_yang_wei_2020}.


Second, we draw inspiration from recent work in neuroscience suggesting that, in addition to TD-error, experiences should also be prioritized in terms of their ``need''. The need term estimates the relevance of each state for future behavior. In particular, it estimates the expected (discounted) number of visits to each other state (or features, in the case of function approximation). The normative reasoning is as follows: if the agent does not expect to return to a particular state, any Q-update pertaining to that state is irrelevant, since the agent would not ``make use'' of what was learned from the replay. Similarly, the more frequently an agent expects to visit a certain state, the more useful are Q-updates pertaining to actions available in that state, since the agent would be able to make use of the corresponding updates multiple times. Note that a low-need experience should be de-prioritized even if is has a large TD-error. Similar arguments can be made for the case of function approximation, substituting states by features where appropriate. In neuroscience, the need term has been shown to play a critical role in dictating which experiences are retrieved from memory \citep{gershman2012successor} and which experiences are replayed during hippocampal replay \citep{mattar_daw_2018}. This latter work, in particular, suggests that prioritizing backups on the basis of both prediction error and the need term can lead to faster learning in biological organisms. \citep{mattar_daw_2018}. Thus, this work motivates our research into whether we can improve Q-learning algorithms by adopting a similar pattern to experience replay algorithms based on TD-error only. 

\section{Computing ``need'' via the Successor Representation}
\label{successor representation}
Our proposal is that the efficiency of experience replay can be increased by considering, in addition to TD-error, the ``need'' of each experience -- namely, experiences with high need should have higher priority for replay. In this section, we provide a formal definition for the need term, and show that it can be computed from the Successor Representation.

The need term is formally defined as the expected (discounted) number of future visits to the state where each experience was encoded. Here, we consider this expectation conditioned on the agent's current state. For example, assume the experience $t = (s_j, a_j, r_j, s_{j+1}, a_{j+1}, p_j)$ is present in the replay buffer, and the agent's current state is $s_i$. By definition, the need term of this transition is: 
\begin{equation}
    Need(s_i, s_j) = \mathbb{E}\Bigg[\sum\limits_{t=0}^\infty \gamma^t \mathds{1}[s_t = s_j]\bigg| s_0 = s_i \Bigg]
    \label{def_need}
\end{equation}
The higher the need of an experience, the more frequently the agent is expected to return to the experience's starting state in the future. Thus the more relevant is that experience with regards to the current state.

In order to compute this term in reinforcement learning, we use the successor representation \citep{dayan_1993} (see also \citep{kulkarni_saeedi_gautam_gershman_2016}). The successor representation estimates the expected discounted future occupancy of every other state, starting from the current state. Denoting the state-to-state transition function as $T$, the SR can be defined as:
\begin{equation}
    \mathbf{M} = (I - \gamma T)^{-1}
    \label{sr_def_dayan}
\end{equation}
 \citep{dayan_1993}. Importantly, in the tabular setting, the elements of this matrix can be shown to correspond exactly to need terms as in Equation \ref{def_need}, where the expectation is taken over both the agent's policy and the environment transitions.

The SR has also been generalized to the case of linear function approximation (e.g., \cite{barreto2018successor}) as well as non-linear function approximation (e.g., \cite{kulkarni_saeedi_gautam_gershman_2016}). For each of these cases, we describe below how the Successor Representation is defined, how it can be learned from experience, and how the need term can be derived from it.


\subsection{Tabular \& linear function approximation}
\label{tabular methods}

While the direct calculation of SR matrix $\mathbf{M}$ using Equation \ref{sr_def_dayan} is intractable because the transition matrix $T$ is often unknown, it can be learned via TD-learning as a special case of Successor Feature (SF) \citep{barreto2018successor}. In the work of successor features, the reward function is decoupled into a linear product of $R(s) = \phi(s)\cdot \mathbf{w}$. The main goal is to learn the SF vector $\psi(s, a)$ of any state-action pair in order to directly 
estimate value function $Q(s, a) = \psi(s, a)\cdot\mathbf{w}$, where $\psi(s, a) = \sum_{s'} M(s, s')\phi(s')$.

Note that when the state feature $\phi(s)$ is one-hot with a dimensionality of exactly $|S|$, learning the SF reduces to the special case of learning the tabular SR matrix $\mathbf{M}$ \citep{barreto2018successor}, since $\psi(s, a) = \sum_{s'} M(s, s')\phi(s') = \mathbf{M}_s$, where $\mathbf{M}_s$ is the row in $\mathbf{M}$ that corresponds to state $s$ with entries that equals to the expected future occupancy of any other states $s'$. Thus, by using a one-hot state representation $\phi(s)$, we are able to learn the tabular SR matrix $\mathbf{M}$ in a vectorized manner using the exact same TD-learning rules for SF \citep{barreto2018successor}. Below we present a definition for the SR and describe the learning rules.

\textbf{Definition: }The definition of the SR matrix $\mathbf{M}$ in tabular case is:

\begin{equation}
\mathbf{M}_{ij} = \mathbb{E}\Bigg[\sum\limits_{t=0}^\infty \gamma^t \mathds{1}[s_t = s_j]\bigg| s_0 = s_i \Bigg]
\end{equation}

\textbf{Learning rule: }We use one-hot features $\phi(s)$ in order to simplify the TD-learning of $\mathbf{M}$. Note that a TD$(\lambda)$ learning has been used (see Appendex \ref{appendix_tabular}).
\begin{equation}
    e_{t+1} \leftarrow \gamma\lambda e_t + \phi(s_t)
\end{equation}
\begin{equation}
    \mathbf{M}_{t+1} \leftarrow  \mathbf{M}_t + \alpha e_{t+1} \Big[\phi(s_t) + \gamma \phi(s_{t+1})M_t - \phi(s_t)\mathbf{M}_t\Big]
\label{sf_td_learning}
\end{equation}
\textbf{Need term: }
The need term is the entry at the $i$th row and $j$th column, which represents the expected count of future visits from state $s_i$ to $s_j$.
\begin{equation}
   Need(s_i, s_j) = \mathbf{M}_{ij}
\end{equation}

\subsection{Nonlinear function approximation}
\label{non-tabular methods}

For larger state spaces than simple tabular ones, non-linear function approximation is used, such that each state $s$ is represented by a feature vector $\phi(s)$ from a deep neural network \cite{kulkarni_saeedi_gautam_gershman_2016}. Similar to successor feature (SF) $\psi(s)$ in section \ref{tabular methods}, we now aim to learn an SR feature vector $\mathbf{m}_{s, a}$.

To get the need term, while we cannot directly extract it as the entry in SR matrix $\mathbf{M}$ since the state space is huge, we can estimate it from $\mathbf{m}_{s, a}$ in Equation \ref{dsr_vector_def} using vector arithmetic.

\textbf{Definition: }The SR feature here is defined as a sum of other state features $\phi(s')$, each being weighted by its expected number of future visits $M(s, s', a)$, which is exactly the need term. 
\begin{equation}
\mathbf{m}_{s, a} = \sum\limits_{s'\in\mathcal{S}}M(s, s', a)\phi(s')
\label{dsr_vector_def}
\end{equation}
\textbf{Learning rule: }We use a similar approach as the TD-learning rule for SF (Equation \ref{sf_td_learning}), except that the feature vector $\mathbf{m}_{s_t, a_t}$ here is from a neural network. 
\begin{equation}
    \mathbf{m}_{s_t, a_t} \leftarrow \mathbf{m}_{s_t, a_t} + \alpha \Big[\phi(s_t) + \gamma\cdot \mathbf{m}_{s_{t+1}, \arg\max_a Q(s_{t+1}, a)} - \mathbf{m}_{s_t, a_t}\Big]
    \label{dsr_learning}
\end{equation}
To learn $m_{s, a}$ which is in vector form outputted from a neural network, we train the network by minimizing the the loss $\ell_u$. Meanwhile, to learn state feature $\phi(s)$ from a neural network, the loss $\ell_g$ is also minimized. 
\begin{equation}
    \ell_g = \left\Vert(s_t - g_{\theta_{\text{sr}}}(\phi_{t})) \right\Vert^2
    \label{l_g}
\end{equation}
\begin{equation}
\begin{aligned}
    \ell_u =  \Vert \big( \phi_{t} + \gamma \cdot u_{\theta_{\text{sr}}}(\phi_{t+1}, \arg \max_a Q(s_{t+1}, a)) - u_{\theta_{\text{sr}}}(\phi_{t}, a_{t}) \big)\Vert^2
\end{aligned}
\label{l_u}
\end{equation}
\textbf{Need term: }Using definition in Equation \ref{dsr_vector_def}, we estimate the need term using vector projection.
\begin{equation}
    Need(s, s') = M(s, s', a) \approx \frac{\mathbf{m}_{s, a} \cdot \phi(s')}{\left\Vert\phi(s')\right\Vert^2}
\end{equation}

\section{Experience replay with SR}
\label{experience replay with sr}

So far we have shown how to learn and implement the need term with successor representation in both tabular and non-tabular cases. In this section, we use the need term to augment two existing algorithms: Prioritized Sweeping (PS) \citep{sutton_barto_2014} and Prioritized Experience Replay (PER) \citep{schaul_quan_antonoglou_silver_2016}. Both of these algorithms prioritize experiences solely on the basis of TD-error. Here, we show that augmenting these prioritization schemes with the need term confers significant improvements to these algorithms.

\subsection{Prioritized sweeping with Successor Representation (PS-SR)}
\label{PS-SR}

We started with Prioritized Sweeping in the Dyna-Q maze game \citep{sutton_barto_2014}. As we have illustrated in section \ref{tabular methods}, the successor representation matrix $\mathbf{M}$ will be learned using the TD($\lambda$) method, and the need terms are then extracted from $\mathbf{M}$ directly to prioritize the experiences in the replay buffer. 

\begin{figure}[ht]
\vskip 0.2in
\begin{center}
\centerline{\includegraphics[width=\columnwidth]{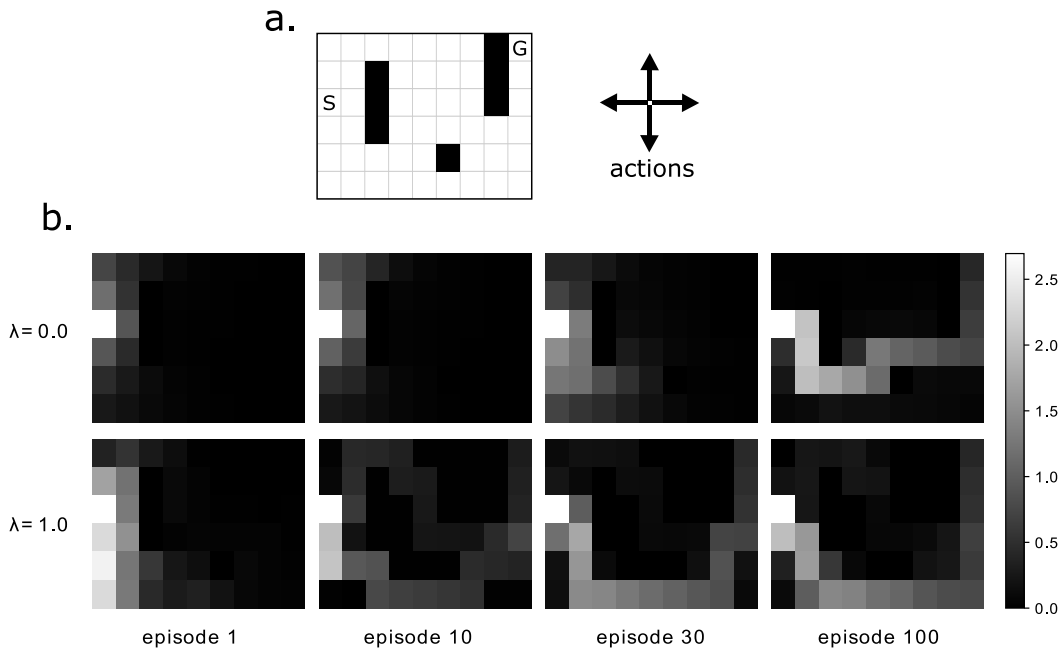}}
\caption{\textbf{a}: The Dyna Maze game board, where S is the start state and G is the goal state. The agent tries to reach the goal state by taking one of four actions in each state. \textbf{b}: Learning the SR matrix $\mathbf{M}$ with the environment shown in \textbf{a}, given $\lambda = 0$ verses $\lambda = 1. $\textbf{Upper row:} The expected number of future visits of every state in the Dyna-Q maze game given the start state as the current state, after learning the SR matrix $\mathbf{M}$ for 1, 10, 30, 100 episodes. The agent is using TD$(\lambda=0)$. The value decays along the optimal path that the agent has found. \textbf{Lower row:} The same learning environment but with TD$(\lambda=1.0)$. It can be seen that it helps the agent to learn the SR matrix faster, as the path appears to be more salient given the same number of episodes learned.}
\label{sr}
\end{center}
\vskip -0.2in
\end{figure}

The effect of using TD$(\lambda)$ eligibility trace is shown in Figure \ref{sr}. Under the simple environment of the Dyna-Q maze game, we compare the learned matrices at the same number of episodes with $\lambda=0$ and $\lambda=1.0$. In $\mathbf{M}$ we take the row corresponding to the start state and reshape it into a heatmap, where cells with lighter color indicates larger values, meaning that these cells are going to be visited more frequently after visiting the start state. Ideally, we want the cells close to the terminal cell to be lighter in fewer episodes, which means that we have learned the SR values fast. The result is that with a larger $\lambda$ value, the agent seems to be able to learn the SR matrix faster that corresponds to the current policy. As in TD-learning, the effect of lambda is to control the amount of bootstrapping. Lower values of lambda lead to more bootstrapping, and thus slower convergence and lower variance. Conversely, higher values of lambda lead to less bootstrapping, and thus faster convergence at the cost of more variance. To strike a balance, we adopt $\lambda=0.5$.

Recall that in the original Prioritized Sweeping algorithm, a priority queue (buffer) is used to keep all the transitions sorted in descending order by their TD-errors \citep{sutton_barto_2014}. Each time the agent takes an action and observes feedback, it pushes them as a transition into the priority queue with a priority that equals to its TD-error. And when the agent samples a transition, the one in the front of the queue with largest priority (TD-error) would be popped out.

In our proposed algorithm, we consider the priority of each transition the product of its TD-error and its need term, instead of TD-error only. When the agent pushes transitions into the buffer, we still make their priorities equal to their TD-errors as in the original PS. However, when the agent samples from the priority queue, it chooses the one transition with the largest product of its TD-error and need term. Equation \ref{PS_SR sampling} shows this sampling scheme that combines both TD-error and the need term:
\begin{equation}
    i \leftarrow \arg\max_i P_i \cdot \mathbf{M}_{S, S_i}, 1 \leq i \leq |PQueue|
    \label{PS_SR sampling}
\end{equation}
The complete algorithm is presented in Algorithm \ref{PSSR}. In comparison to the original Prioritized Sweeping algorithm, our PSSR algorithm have two additional steps: first, learning the eligibility trace and the successor representation matrix; second, calculating the product of TD-error and the need term when sampling. The rest of the algorithm is exactly the same as the original Prioritized Sweeping. 

\begin{figure}[ht]
\vskip 0.2in
\begin{center}
\centerline{\includegraphics[width=1.0\columnwidth]{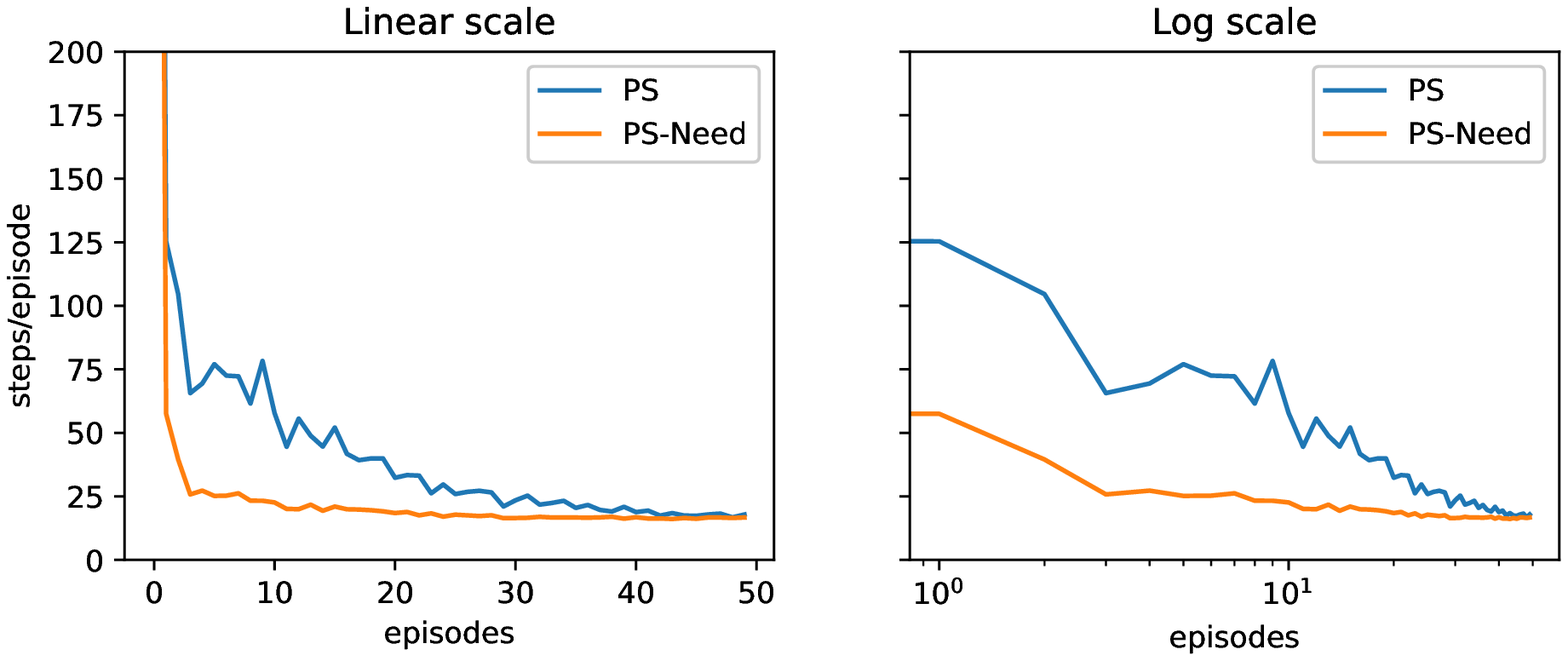}}
\caption{\textbf{Left:} Average number of steps per episode with regard to the number of training episodes. By additionally prioritizing all transitions with the need term, the prioritized sweeping algorithm was able to converge to a shorter path through the Dyna-Q maze with fewer steps per episode. The steps per episode was an average over 50 repetitions. \textbf{Right:} The same figure in log scale.}
\label{ps}
\end{center}
\vskip -0.2in
\end{figure}

We experimented our new algorithm in the Dyna Maze game which was also used by PS\citep{sutton_barto_2014}. Figure \ref{ps} shows that it improves the performance of Prioritized Sweeping, given the example of the Dyna-Q maze. We followed the original experiment settings and an identical structure of the maze board as in Figure \ref{sr}. The successor representation and need term were randomly initialized and learned online. Prioritizing by the need term in addition to just the TD-errors helps the agent find the optimal path faster. While PS (blue curve) takes almost 50 episodes to converge to the shortest-path policy, our proposed algorithm PS-SR (orange curve) reached the optimal policy in much fewer steps. More detail of experimental settings is addressed in Appendix \ref{exp_dyna}.

\subsection{Tabular Prioritized Experience Replay with Successor Representation (vanilla PER-SR)}
\label{vanilla PER-SR}

As we have shown that the need term helps tabular methods (prioritized sweeping) to learn faster, we are then motivated to extend this idea to non-tabular methods such as Prioritized Experience Replay (PER). However, since PER is implemented alongside DQN -- i.e., using a neural network --, this implementation ends up obscuring the inner workings of the sampling algorithm. Thus, before deploying our approach with neural networks, we implemented a simplified version of PER with no networks, by using a trivial function approximation whereby each state is encoded as one-hot feature vectors on an orthonormal basis. The details of the experimental setup are included in Appendix \ref{appendix_cliffwalk}.

\begin{figure}[ht]
\vskip 0.2in
\begin{center}
\centerline{\includegraphics[width=1.0\columnwidth]{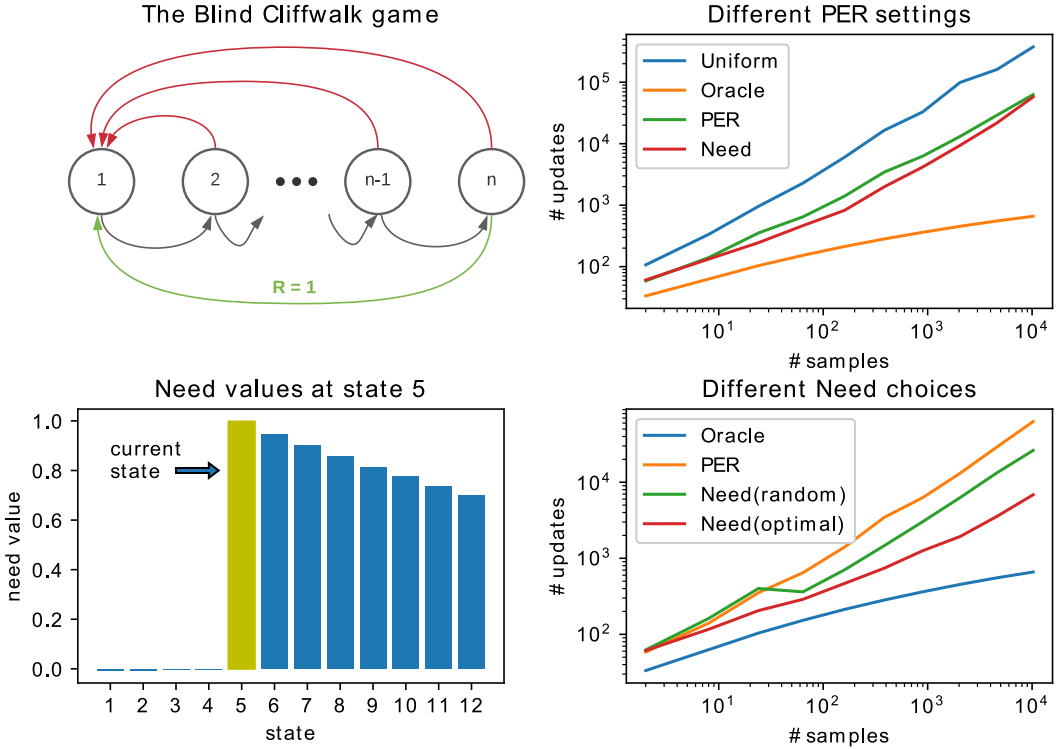}}
\caption{Vanilla PER without neural networks in the Cliffwalk game. \textbf{First}: The Cliffwalk example in \cite{schaul_quan_antonoglou_silver_2016}. \textbf{Second}: Reproduce study in \cite{brittain_bertram_yang_wei_2020} and show that prioritized experience replay with the need terms performs better. \textbf{Third}: The need values (expected count of future visits) of all states at current state 5. \textbf{Fourth}: Comparing different implementations of the need term, which shows that even need terms of random actions can outperform PER, and more improvements can be made if we carefully design a learning rule for the need terms. }
\label{vanilladqn}
\end{center}
\vskip -0.2in
\end{figure}

We used the same experiment settings as use by \cite{schaul_quan_antonoglou_silver_2016}, with the Blind Cliffwalk environment shown in the first plot of Figure \ref{vanilladqn}. We varied the number of states $n$ from 3 to 13. We compared the performances of multiple algorithms, as explained in Appendix \ref{appendix_cliffwalk}. In order to implement the vanilla PER algorithm and the need term without neural networks, we used tabular SR (Section \ref{tabular methods}) and one-hot vector state representation. The full algorithm is in Algorithm \ref{vanilla PER algo}.

To use the need term in sampling, we changed the calculation of the sampling probabilities to Equation \ref{prop_sample_eq}. The need terms were successfully learned by the agent, given the example shown in the third plot in Figure \ref{vanilladqn}, where the agent correctly learned the need terms of all other states when the current state was the fifth one. While the states before the current state had low need values, the ones after it had much higher ones and decayed as the distance became farther. Intuitively, high need values on the right side made the agent more likely to sample them and learn their Q-values, which helped the agent take better actions when it really reaches those states on the right side afterwards.
\begin{equation}
\begin{split}
    P(j) &= \frac{(p_j \cdot Need(s_t, s_j))^\alpha}{\sum_i (p_i \cdot Need(s_t, s_i))^\alpha}\\
    &= \frac{(p_j \cdot M(s_t, s_j))^\alpha}{\sum_i (p_i \cdot M(s_t, s_j))^\alpha}
\end{split}
\label{prop_sample_eq}
\end{equation}
As we have successfully obtained a similar result that PER was exponentially faster to converge (green), we additionally tested our PER variant with the need term and showed that PER with the need term (red) was taking even fewer Q-updates (Figure \ref{vanilladqn}). Meanwhile, the result showed that surprisingly even the random scheme outperformed vanilla PER (the last plot of Figure \ref{vanilladqn}). The optimal need terms then improved even more, getting closer to the lower bound of the oracle one. Thus, it is of our particular interest to design an implementation of the need terms that would lie between performances of the random and optimal schemes. 

\subsection{Prioritized Experience Replay with Deep Successor Representation (PER-SR)}
\label{PER-SR}

Motivated by the success in vanilla PER, we finally integrated the need term into PER using deep neural networks. We learned the SR and the need term using the rules explained in section \ref{non-tabular methods}. We have shown earlier in PS-SR (section \ref{PS-SR}) and vanilla PER (section \ref{vanilla PER-SR}) that the need term could help the agent replay more important experiences by changing the sampling probabilities.  However, this was achieved by calculating the need terms of every single experience in the buffer w.r.t. the current state every time we need to replay some experiences. This could be very computationally inefficient since the buffer size and state space might be too large in complex environments (e.g. Atari games). 

To get around this issue, instead of modulating the sampling probability, we used the need term to modulate the magnitudes of the Q-updates directly. Each minibatch of transitions being sampled was still based on the probabilities calculated solely from TD-errors as in \cite{schaul_quan_antonoglou_silver_2016}. However, to replay each transition and perform the Q-update, we adjusted this Q-update by multiplying the need value of this transition. So among all important transitions measured by high TD-errors, we managed to further choose those that are highly relevant in the near future, indicated by high need values. Intuitively, this alternative was based on the idea that either increasing the sampling probability or enlarging the magnitude of a backup could result in larger expected Q-update (see also \ref{discussion}). Hence, we can derive our new sampling scheme of both TD-error and the need term, from directly sampling using TD-error only, and resize the magnitude of the Q-backup with the need term $Need(s_k)$.

\begin{figure}[ht]
\vskip 0.2in
\begin{center}
\centerline{\includegraphics[width=\columnwidth]{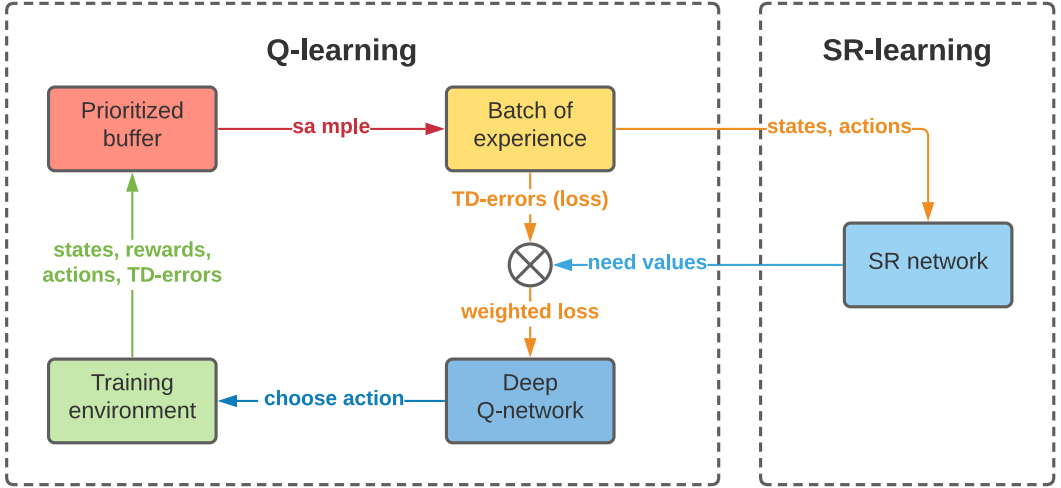}}
\caption{The overall structure of PER with deep SR, as implemented by Algorithm \ref{PERSR algo}. The calculation of the need terms from the SR network is addressed in Section \ref{non-tabular methods}, and the structure of the SR network (shown on the right side) is demonstrated in Figure \ref{fig:DSR}.}
\label{PERSR}
\end{center}
\vskip -0.2in
\end{figure}

When implementing our approach in deep neural networks, we also found that the the estimated SR vectors $m_{s, a}$ were inaccurate early in training and sometimes led to negative values. We therefore applied a constant offset to all need values so that the smallest need was zero. As shown in equation \ref{need_anneal} below, we calculated the need terms of all $k$ transitions in the minibatch and subtract the lowest. Other approaches to mitigate this issue were also attempted (see \ref{discussion}).
\begin{equation}
\begin{aligned}
    Need(s') = M(s, s', a) - \min\big(0, \min_i M(s, s_i, a)\big), i\in [1, k]
\end{aligned}
\label{need_anneal}
\end{equation}
The full PER-SR algorithm is presented in Algorithm \ref{PERSR algo}. Our main modifications to the original PER would be adding the training step for the deep SR network, and calculating the need terms for the experiences being sampled, as shown in the ``SR-learning" area in Figure \ref{PERSR}. All other parts of the PER algorithm remains unchanged. We experimented this algorithm in the Atari benchmarks in section \ref{atari_experiments}.

\section{Atari experiments}
\label{atari_experiments}

With the details of implementation of PER-SR presented above, we experimented it in a selected set of Atari games. This selected set was considerably small, however, on the contrary to 49 games which the original work of PER had used. Nevertheless, the result was promising and strongly suggested a performance improvement after using the need term.

The baseline algorithm we used was the double DQN as used in the PSER paper by \citep{brittain_bertram_yang_wei_2020}. We also used the same Dopamine code base \citep{castro2019dopamine} as they did. The parameters we used for experience replay were exactly the same as the ones in the PSER paper \citep{brittain_bertram_yang_wei_2020}, while the other parameters for DQN were the same as in the Nature paper \citep{DBLP:journals/nature/MnihKSRVBGRFOPB15}. The learning rate $\alpha$ for the deep SR network was by default the same as the Q-network. 

\begin{figure}[ht]
\vskip 0.2in
\begin{center}
\centerline{\includegraphics[width=\columnwidth]{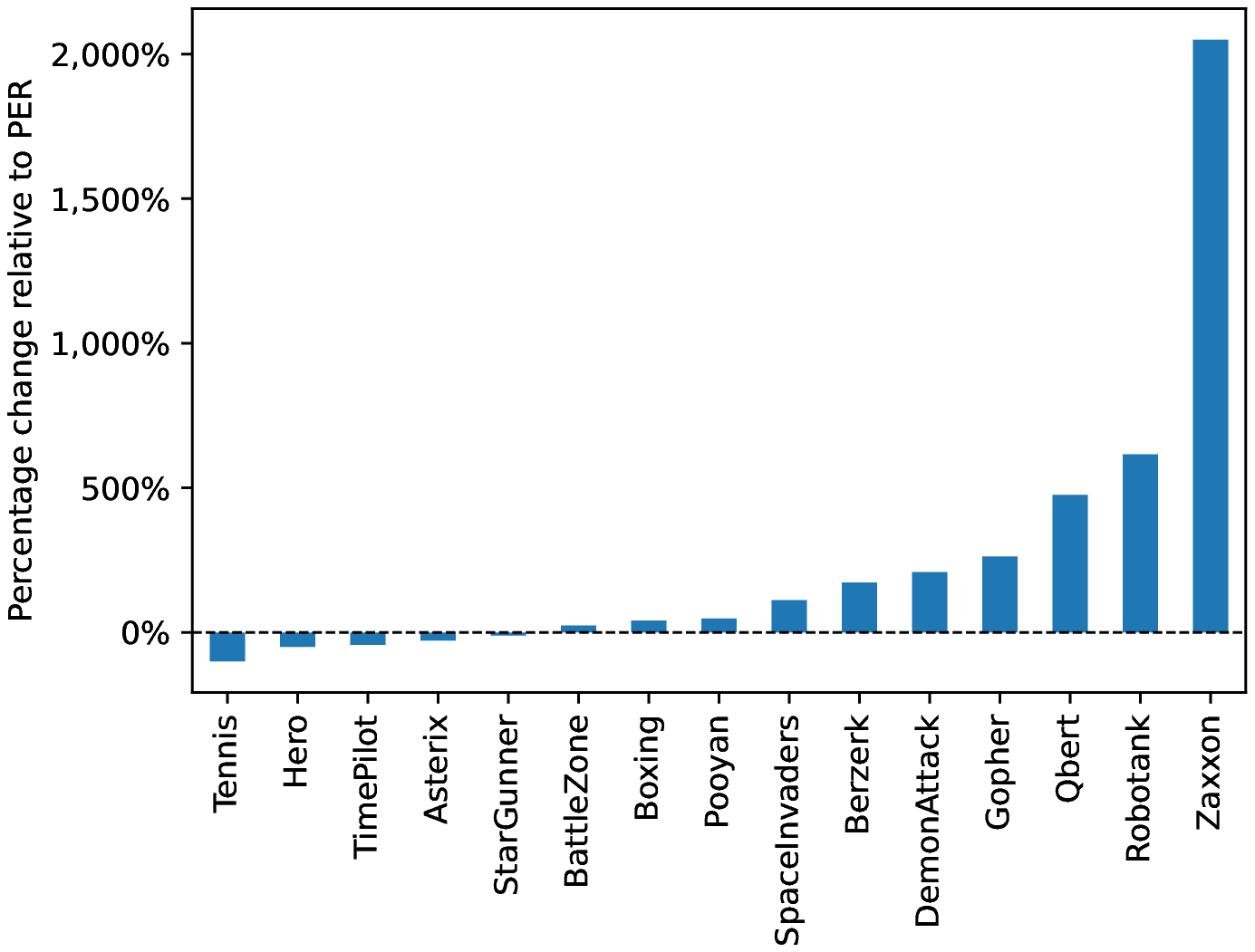}}
\caption{The relative performance on a set of selected games of PER-SR with regard to original PER. The games were selected from the 49 games in the original PER paper. The dotted line means the same performance as original PER. Any bar that goes above the line indicates a performance improvement of PER-SR over PER.}
\label{perfcomp}
\end{center}
\vskip -0.2in
\end{figure}

We produced our results by training the algorithms on all games for 200 iterations. Each iteration consisted of 250K observations, which corresponded to 1 million frames as the games are using frame skipping with a factor of 4 \citep{DBLP:journals/nature/MnihKSRVBGRFOPB15}. After each one training iteration finished, we evaluated the agent for 125K observations, which corresponded to 30 minutes of game play. The average game score in all episodes of the evaluation pass of one single iteration was taken, so in the end we had 200 average returns for each agent in a game. The final measurement of the agent's performance was the performance of the best policy, which was simply the maximum of the 200 averages returns. With this performance metric, the summarized performance comparison of PER and PER-SR was shown in Figure \ref{perfcomp}. The vertical axis measures the percent change of PER-SR's raw score with regard to PER's raw score, where any value that is larger than zero (above the dotted line) indicates performance improvement in the game. Overall, PER-SR achieved better performance in most of the games, and there were significant performance increase in 3 game (Berzerk, Qbert, Robotank). 

The learning curves of all games in detail are shown in Figure \ref{atari}. We observed that while PER-SR suggested improved scores over the original PER, it learned comparatively slower and achieves lower scores in the early stage of learning. For example, while PER-SR ultimately reached higher scores in the game SpaceInvaders, its score improved slowly compared to PER which quickly achieved a score of 600 just in a few million frames. This slower learning process was probably due to the fact that PER-SR needed a good SR network to provide an accurate estimation of the need term which could improve the prioritization of experience replay. It might take a while at first for the agent to explore the environment and train the deep successor representation. Thus, the need values were probably not that accurate in the beginning, which did not help prioritize the most relevant experience as it later might do.

Furthermore, it also showed that while PER seemed to learn a good policy very fast at first, in the end the performance drops. For example, in the game Boxing (Figure \ref{atari}), while PER achieved a score of 50 much faster than PER-SR in just a few million steps, the score dropped drastically to as low as -40 afterwards. This is probably because PER had overfitted the Q-fucntion very quickly, since PER chooses frequently chooses those transitions with high TD-errors over and over again. The authors of PER addressed this problem by incorporating an importance sampling weight to downsample those transitions with high TD-errors \citep{schaul_quan_antonoglou_silver_2016}. The overfitting problem might still persist even after applying these weights. However, in our new approach, by scaling the updates with the need terms, we seemed to mitigate this effect, as in most of the games PER-SR continued to achieve higher scores after training for 200 million steps without going back to a very low score. We speculate that by rescaling the Q-updates of the transitions using the need terms, the agent will become less likely to overfit the data with the highest TD-errors. 

\section{Discussion}
\label{discussion}

We have shown that our proposed algorithm PER-SR exhibited significant performance increase across different game environments in the Atari benchmarks. Meanwhile, through implementing this new algorithm by experimenting on different variants, we observed some important issues which could possibly motivate future research, in order to understand how the idea of successor representation can lead to even better prioritization of experience replay and result in more promising experiment performance. 

\textbf{Environments:} While our new algorithm PER-SR showed an overall performance increase in our selection of Atari games, we speculate that the extent to which the game scores were improved depended on the nature of each game itself. We observed that games with the following traits tended to score higher with SR: first, the states at different time steps appeared very similar; second, same (or similar) states tended to frequently recur through many episodes. It can be reasonably hypothesized that these traits can make the agent learn the deep SR network faster, which then provides an accurate estimation of the need term that helps prioritize the most relevant experiences. Thus, we can possibly expect performance improvements by applying PER-SR to other tasks with state space that reflect these traits in future research. 

\textbf{Magnitude vs. probability: }In our implementation of PER-SR, we still sampled each minibatch with a probability distribution based on the priority (TD-error) of each transition, same as the original PER. We then resized the magnitude of the backup of each transition by multiplying it with its need value, which turned out to help PER learn faster. However, it is yet unknown whether the same or better improvement can be achieved by directly changing the sample probability based on the product of TD-error and need term of each transition. Currently, this probability-based approach has yet not been used, because it was extremely time consuming to calculate the need terms for all transitions based on the new current state every time we want to sample a batch, especially when the replay buffer had grown considerably large with millions of transitions after learning for many episodes. Nevertheless, it is possible that future research may provide solutions to this problem, and finally show the effects of this alternate probability-based approach. 

\textbf{Forward vs. backward relevance: }Our PER-SR algorithm used the need term to prioritize those transitions that are going to be frequently visited afterwards. Often, those experiences that happen right after the current state will have high need values. However, it is unknown to us if we can also prioritize the experiences that are likely to frequently lead to the current state. It has been shown by \cite{sutton_barto_2014} that the sweeping across all transitions predicted to lead to the current state is essential for the faster convergence of the prioritized sweeping algorithm (PS) over Dyna-Q. Hence, it motivates future research about whether a similar approach of experience replay by the relevance before the current state would also improve PER. More variants of the successor representation may also be proposed as a way to not only account for the future occupancy but also for the relevance in the past. 

\textbf{Negative relevance: }As mentioned earlier in section \ref{PER-SR}, due to the inaccuracy of learning the SR vectors $m_{s, a}$ and the state features $\phi(s)$ with the deep SR network, the need term calculated from the SR vectors might be negative. In our paper, this was addressed by subtracting the minimum need term in a sampled minibatch. However, there can be other approaches. For example, when the need terms of some experiences in the minibatch are negative, we can simply choose not to rescale the magnitudes of the Q-updates, which simply reduces to normal prioritized experience replay by TD-errors. 

\section{Conclusion}

This paper introduces a novel prioritization metric that combines both TD-error and the need term, which measures the relevance of each experience using the successor representation. We studied both tabular and non-tabular methods, implemented our extension to both Prioritized Sweeping (PS) and Prioritized Experience Replay (PER) algorithms, and found that our proposed algorithms led to significantly better performance in the Dyna Maze and Atari game benchmarks. Our work suggests a promising direction for improving experience replay in future works and motivates further research in topics including the issues we discussed so far. 

\subsubsection*{Acknowledgments}
We thank Nautilus, a kubernetes cluster at UCSD, which provided us great computing resources which made our experiments possible. 

\bibliography{main_paper}
\bibliographystyle{main_paper}

\clearpage
\appendix

\section{Successor Representation (SR)}
\label{appendix_tabular}

\subsection{Tabular \& linear function approximation}
\textbf{Definition: }As mentioned earlier, the tabular successor representation is defined as $M = (I - \gamma T)^{-1}$ by \cite{dayan_1993}, and learned using the same rules for the successor features (SF) \citep{barreto2018successor}. $\mathbf{M}$ is a  matrix where the entry $M_{ij}$ at row $i$ and column $j$ represents the expected number of visits to state $s_j$ from the current state $s_i$. $T$ is the state transition matrix where the entry $T_{ij}$ at row $i$ and column $j$ means the probability that the next state is $s_j$ given the current state $s_i$. From the definitions of the need term and the tabular SR, we can see that the need term $Need(s_i, t)$ in equation \ref{def_need} can be directly taken from $\mathbf{M}$:
\begin{equation*}
    Need(s_i, t) = \mathbf{M}_{ij} =  \mathbb{E}\Bigg[\sum\limits_{t=0}^\infty \gamma^t \mathds{1}[s_t = s_j]\bigg| s_0 = s_i  \Bigg]
\end{equation*}
\textbf{Learning rule: }The SR matrix $\mathbf{M}$ cannot be learned from the transition matrix $T$ directly, because $T$ cannot be easily obtained as the probability distribution of state transitions remain unknown unless the agent has run for a significant period time in a new environment. Our solution is to instead use TD($\lambda$), a temporal-difference learning method based on the eligibility trace \citep{sutton_barto_2014}. 

Therefore, we aim to learn the successor representation matrix $\mathbf{M}$ using TD($\lambda$) learning. The matrix $\mathbf{M}$ is a square matrix of rank $n$, where $n$ is the total number of states in the environment, and each state is represented as a one-hot vector $\phi(s)$. Each row is a weighted sum of all $n$ feature vectors: 
\begin{equation}
    \mathbf{M}_i = \mathbf{M}_{i1}\phi(s_1) + \mathbf{M}_{i2}\phi(s_2) + \cdots + \mathbf{M}_{in}\phi(s_n)
\end{equation}
The complete TD($\lambda$) equations are listed as below. Note that we initialize the eligibility trace vector $e$ as zeros, and the SR matrix $\mathbf{M}$ using $(I - \gamma T)^{-1}$, where $T$ is initialized assuming that the agent behaves random actions at each state. The feature vector $\phi(s)$ is a row vector, whereas the eligibility trace vector $e$ is a column vector. 
\begin{equation}
    e_{t+1} \leftarrow \gamma\lambda e_t + \phi(s_t)
\end{equation}
\begin{equation}
\begin{aligned}
    \mathbf{M}_{t+1} \leftarrow \mathbf{M}_t + \alpha e_{t+1} \Big[\phi(s_t) + \gamma \phi(s_{t+1})\mathbf{M}_t - \phi(s_t)\mathbf{M}_t\Big]
\end{aligned}
\end{equation}
\textbf{Need term: }As shown in Equation \ref{need_equation_tabular}, the need term can be directly obtained from entries in the learned SR matrix $\mathbf{M}$. We will incorporate it into Prioritized Sweeping (PS), an existing experience replay technique by TD-errors based on the tabular Dyna-Q algorithm. We will show the new algorithm that uses the need term, and also the effect of using the eligibility trace to the learning speed of successor representation in section \ref{PS-SR}.
\begin{equation}
   Need(s_i, s_j) = [\mathbf{M}_{t+1}]_{ij}
   \label{need_equation_tabular}
\end{equation}
\subsection{Nonlinear function approximation}
\label{appendix_nontabular}
For non-tabular (deep learning) methods, the definition of SR, the learning rule for SR, and the calculation of the need term with SR are all different. The reason is that in the tasks which we solve using non-tabular methods, the state space is either significantly large, or continuous, such that the state representation of one-hot vectors cannot be used anymore. We are thus not able to define the successor representation as a matrix $\mathbf{M}$ which we used in tabular cases. 

\begin{figure}[ht]
\vskip 0.2in
\begin{center}
\centerline{\includegraphics[width=0.8\columnwidth]{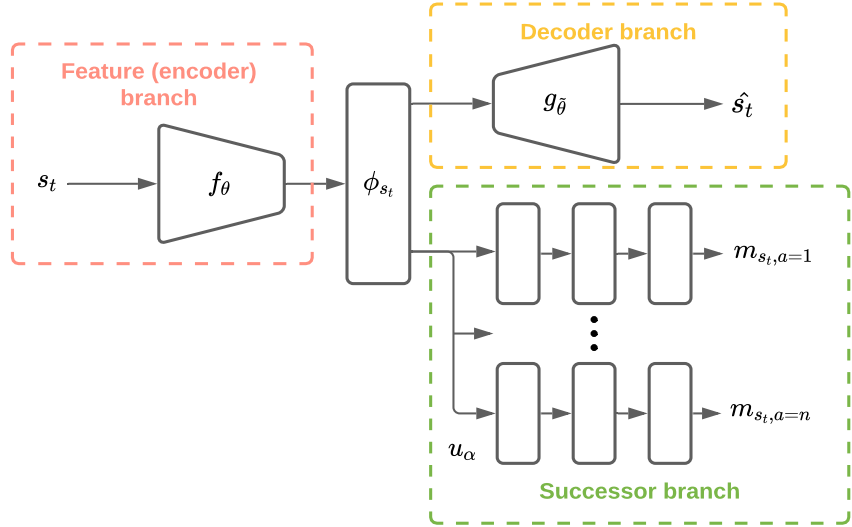}}
\caption{Our simplified version of the architecture of the Deep Successor Representation network in \cite{kulkarni_saeedi_gautam_gershman_2016}. Note that it is almost exactly the same as the original one but without the layers to learn the reward function. }
\label{fig:DSR}
\end{center}
\vskip -0.2in
\end{figure}

\textbf{Definition: }We alternatively define the successor representation as a vector, as opposed to a matrix in tabular cases. The SR vector, denoted as $m_{s, a}$, is a weighted sum of all other state feature vectors $\phi(s')$, each being multiplied by the expected count of future visits represented by $M(s, s', a)$. It is clear that the need term we want is exactly $M(s, s', a)$. The equations of this definition for the SR are as below \citep{kulkarni_saeedi_gautam_gershman_2016}:
\begin{equation}
M(s, s', a) = \mathbb{E}\Bigg[\sum\limits_{t=0}^\infty \gamma^t \mathds{1}[s_t = s']\bigg| s_0 = s, a_0 = a  \Bigg]
\end{equation}
\begin{equation}
\mathbf{m}_{s, a} = \sum\limits_{s'\in\mathcal{S}}M(s, s', a)\phi(s')
\end{equation}
\textbf{Learning rule: }Given the definition of the SR, we now present its learning rules. We aim to learn the SR vector $m_{s, a}$, assuming that we already have a representation vector $\phi(s)$ for each state $s$. We still use a temporal-difference method. We mentioned in equation \ref{dsr_vector_def} that the SR vector is a weighted sum of feature vectors by future occupancy. Based on this, when we take action $a_t$ at state $s_t$, the SR vector $m_{s_t, a_t}$ should equal to the SR vector $m_{s_{t+1}, a_{t+1}}$ plus the feature vector $\phi(s_t)$ of state $s_t$ itself, because the only difference between the two SR vectors is that $m_{s_t, a_t}$ should have one more occurrence of $\phi(s_t)$. We illustrate this learning rule in the below equation \ref{dsr_learning}.
\begin{equation}
\begin{aligned}
    \mathbf{m}_{s_t, a_t} \leftarrow \mathbf{m}_{s_t, a_t} + \alpha \Big[\phi(s_t) + \gamma\cdot \mathbf{m}_{s_{t+1}, \arg \max_a Q(s_{t+1}, a)} - \mathbf{m}_{s_t, a_t}\Big]
    \label{dsr_learning}
\end{aligned}
\end{equation}
To implement an SR that can be learned with the rule explained above, we incorporate the Deep Successor Representation network (DSR) by \cite{kulkarni_saeedi_gautam_gershman_2016}. To learn the SR with the above rule we need to train the neural network in two tasks: first, learning the feature $\phi(s)$ using a autoencoder; second, learning the SR vector $m_{s, a}$. We follow the notation in the DSR, which consists of three main network branches: a feature branch $f_\theta$ which takes in raw images $s_t$ and returns features $\phi_t$; a successor branch $u_\theta$ which computes $m_{s_t, a}$ given feature $\phi_t$ and $a$; and a decoder branch which returns the reconstructed image $\hat{s_t}$ from feature $\phi_t$, as shown in Figure \ref{fig:DSR}. The loss function thus consists a autoencoder loss $\ell_g$, and a SR loss $\ell_u$, as presented in equation \ref{l_g} and equation \ref{l_u}.
\begin{equation}
    \ell_g = \left\Vert(s_t - g_{\theta_{\text{sr}}}(\phi_{t})) \right\Vert^2
    \label{l_g}
\end{equation}
\begin{equation}
\begin{aligned}
    \ell_u =  \Vert \big( \phi_{t} + \gamma \cdot u_{\theta_{\text{sr}}}(\phi_{t+1}, \arg \max_a Q(s_{t+1}, a)) - u_{\theta_{\text{sr}}}(\phi_{t}, a_{t}) \big)\Vert^2
\end{aligned}
\label{l_u}
\end{equation}

\textbf{Need term: }With the learning rule and network implementation of the SR vector presented, we are ready to derive the need term from them. While we mentioned earlier that the need term comes directly from $M(s, s', a)$, it is however not directly learned in the DSR network \citep{kulkarni_saeedi_gautam_gershman_2016}. Instead, we estimate $M(s, s', a)$ backwards from $m_{s, a}$ using vector projection, since we know that $m_{s, a}$ estimates a weighted sum of feature vectors of all other states, as shown in equation \ref{dsr_vector_def}. By using the magnitude of the vector projection of $M(s, s', a)$ in the direction of feature vector $\phi(s')$, we estimate $M(s, s', a)$ as the expected number of future visits to $s'$ given current state-action pair $s, a$. 
\begin{equation}
    Need(s, s') = M(s, s', a) \approx \frac{\mathbf{m}_{s, a} \cdot \phi(s')}{\left\Vert\phi(s')\right\Vert^2}
\end{equation}
With the way to learn the successor representation in the non-tabular case and the derivation of the need term from the SR, we are now able to experiment it in prioritized experience replay (PER), an existing prioritization technique based on DQN. We will show in section \ref{PER-SR} the implementation details of how to add the need term into the original sampling process by TD-error, and finally in section \ref{atari_experiments} we will present the experiment results of our improved version of PER in Atari benchmarks.

\section{Experimental details}
\subsection{Dyna Maze}
\label{exp_dyna}

\begin{figure}[ht]
\vskip 0.2in
\begin{center}
\centerline{\includegraphics[width=0.6\columnwidth]{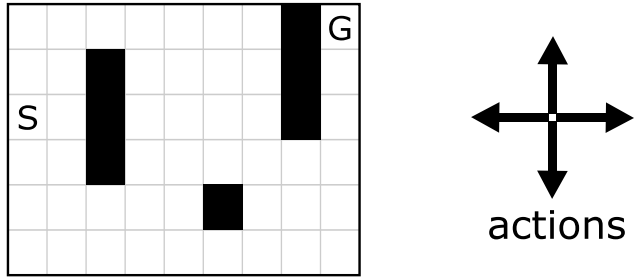}}
\caption{The Dyna Maze environment for the prioritized sweeping algorithm used by \cite{sutton_barto_2014}}.
\label{fig:dyna_maze}
\end{center}
\vskip -0.2in
\end{figure}

For the Dyna Maze experiment (Section \ref{PS-SR}), we used the exact same experiment setting as used by \cite{sutton_barto_2014}. The maze consists of nine columns and six rows, with a start state and a goal state. The goal of the agent is to reach the goal state from the start state in the shortest path possible, given four actions (up, down, left, right) available at each state (square), shown in Figure \ref{fig:dyna_maze}. Note that while all four actions are available at all states, some actions corresponds to invalid moves (for example, going out of the board, or hitting against the wall states in black color). These invalid moves will leave the current state of the agent unchanged. Since the maze is a simple tabular environment, we represented each state by a distinct integer, calculated by $n\_cols * row\_index + col\_index$. For example, the start state $S$ was represented as $9 * 2 + 0 = 18$. 

The reward function was constructed such that every state-action pair, except for the one that lead to the goal state $G$, had a reward $r = 0$. If the agent arrives at the goal state, then it receives a random reward generated by $r \sim N(1, 0.1)$, a normal distribution with a mean of 1 and a standard deviation of 0.1. This was meant to keep the agent learning even if it was close to the optimal policy.

We implemented the prioritized sweeping (PS) algorithm using the exact same implementation by \cite{sutton_barto_2014}. For our new algorithm PS-SR that additionally used the need term in its sampling process, we only added code for learning the tabular need term (Section \ref{tabular methods}) and using it in the sampling process. Apart from them, the new algorithm (shown in Algorithm \ref{PSSR}) was exactly the same as the PS algorithm. The successor representation was defined as a square matrix $\mathbf{M}$ of size $|S|\times|S|$, where $|S|$ was the size of the state space. To initialize and learn $\mathbf{M}$, we use the initialization by $M = (I - \gamma T)^{-1}$ and the TD($\lambda$) learning rule, as explained in Appendix \ref{appendix_tabular}. Note that the transition matrix $T$ was initialized such that it corresponds to a random policy. Each entry $T_{ij}$ was calculated assuming that the four actions each had 0.25 probability to be taken at state $s_i$.

\begin{figure}[ht]
\vskip 0.2in
\begin{center}
\centerline{\includegraphics[width=1.0\columnwidth]{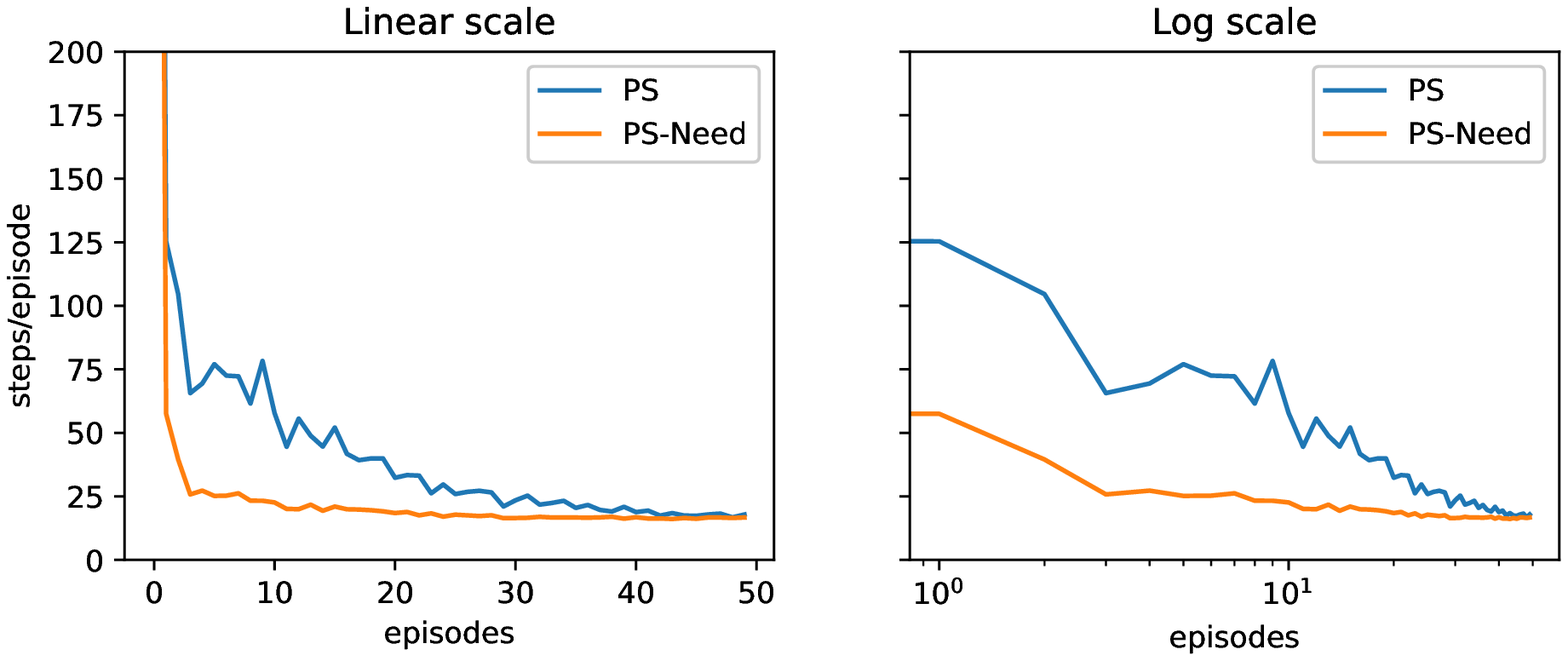}}
\caption{\textbf{Left:} Average number of steps per episode with regard to the number of training episodes. By additionally prioritizing all transitions with the need term, the prioritized sweeping algorithm was able to converge to a shorter path through the Dyna-Q maze with fewer steps per episode. The steps per episode was an average over 50 repetitions. \textbf{Right:} The same figure in log scale.}
\end{center}
\vskip -0.2in

\label{appendix_ps_sr}
\end{figure}

For PS-SR, the replay memory was structured in the same way as the old PS memory, which was a priority queue ordered by the prediction error (TD-error) of each experience (state-action pair) in the replay memory. For example, an experience $(S_i, A_i)$ was ordered by its TD-error $P_i$. However, in PS-SR, instead of directly pop out the first one in the priority queue that has the largest priority, for each experience we calculated the product of its priority and its need term with regard to the current state, noted by $S$. Finally, the experience with the largest product would be sampled and popped out of the memory. This process is illustrated in Equation \ref{appendix_PS_SR_sampling}. 
\begin{equation}
    i \leftarrow \arg\max_i P_i \cdot M_{S, S_i}, 1 \leq i \leq |PQueue|
    \label{appendix_PS_SR_sampling}
\end{equation}
The evaluation metric we used for PS and PS-SR is the length of the episode, since the goal is to make the agent reach the goal state in the shortest path. The parameters were exactly the same as those used in PS by \cite{sutton_barto_2014}. For each step the agent takes, 5 experiences would be sampled from the memory. In order to reduce the variance of the experiment, the experiment consisted of 50 independent trials. In each trial, the agent for both PS and PS-SR would run for 50 episodes and the lengths of all these episodes were recorded. In the end, for each algorithm, we calculated the mean episodic length at each episode number ranging from 1 to 50. The result has been plotted in Figure \ref{appendix_ps_sr}. 

\subsection{Blind Cliffwalk}
\label{appendix_cliffwalk}

We experimented the vanilla PER and multiple other algorithms in the Blind Cliffwalk game as presented in \cite{schaul_quan_antonoglou_silver_2016} and \cite{brittain_bertram_yang_wei_2020}, in order to investigate the extent to which the Need term can help PER under the simple circumstance without any neural networks. The environment is shown in the first plot of Figure \ref{vanilladqn}, where the goal was to go from the start state at the leftmost side and reach the goal state at the rightmost side. At each state, the correct action for the agent was to go to the next state on the right side. If it failed to do so, it ``falled'' down the cliff and returned to the start state. To determine how fast an agent learns an optimal policy, we measured the number of Q-updates needed for the agent to converge to the optimal Q-function. This convergence was defined as a MSE error between the Q-value estimates and the ground-truth Q-values below a threshold of $10^{-3}$. The ground-truth Q-function was easily calculated in advance from the observation that the only the goal state had a reward of 1 and the best action should always be going right for all states. 

We reproduced the experiment in \cite{schaul_quan_antonoglou_silver_2016} that compared performance of three algorithms: uniform, oracle, and PER. To investigate the effect of the need term, we additionally tested our PER variant with the need term, noted as "Need" in the plot. In order to determine only the computational and sampling performance of the algorithms, it is important that the samples were pushed to the buffer before the agent started running as done in \cite{brittain_bertram_yang_wei_2020}. The samples were generated by exhaustively taking all $2^n$ random sequences of actions \citep{schaul_quan_antonoglou_silver_2016}, and no more samples were pushed to the buffer when the algorithms were running. The four algorithms were constructed as below, and the results were shown in the second plot of Figure \ref{vanilladqn}:

\textbf{Uniform: }The uniform one represented a uniform memory replay scheme, which was identical to the one used in original DQN \citep{DBLP:journals/nature/MnihKSRVBGRFOPB15}. To implement this, we simply used PER and set the prioritization exponent $\alpha$ to zero.

\textbf{Oracle: } Each time we choose a state-action pair to perform the Q-update, we iterated through all possible state-action pairs and calculated their amounts of Q-updates if they were really taken. The Q-update that would cause the largest amount of Q-update would be chosen and performed. This algorithm used our knowledge in hindsight to make the algorithm converge to the ground-truth Q-values in the fastest way, which represented a lower-bound of the number of Q-updates. 

\textbf{PER: } Used the same settings as in \cite{schaul_quan_antonoglou_silver_2016}. The sampling exponent $\alpha = 0.6$. Instead of approximating the Q-function using deep neural networks and find its gradient w.r.t. the parameters of the network using back propagation as in DQN by \cite{DBLP:journals/nature/MnihKSRVBGRFOPB15}, we represented the Q-function in the ``vanilla PER'' as $Q(s, a) = \theta \cdot \phi(s, a)$, where $\phi(s, a)$ was the one-hot vector representation of discrete state-action pairs and $w$ is a vector consisting of parameters of this Q-function. Thus, the analytical solution of the Q function w.r.t. its parameters $w$ could be directly found as $\nabla_\theta Q(s, a) = \phi(s, a)$, which simply reduced to the one-hot vector that represents the pair $(s, a)$. It can also be easily seen that the entries in the parameters vector $\theta$ are just the Q-value estimates of the state-action pairs. The other parts of PER, including the setup of the replay buffer, exactly followed the algorithm in \cite{schaul_quan_antonoglou_silver_2016}.

\textbf{Need: } For PER with Need, we used the tabular SR, illustrated in section \ref{tabular methods}. As illustrated in Section \ref{PS-SR} the effect of TD($\lambda$) learning the SR faster, we use a $\lambda = 0.95$. The SR matrix $\mathbf{M}$ is a $|S|\times|S|$ square matrix, where each row and column corresponds to each state. To calculate the need term, we directly extract the entry in the SR matrix $\mathbf{M}$ at the column and row that corresponds to the states, as mentioned in Appendix \ref{appendix_tabular}. The need term was directly used to calculate the sampling probabilities of every single experience in the replay buffer: we calculated the priority of each experience as the product of its TD-error and need term, and then used this new priority to compute the sampling probability with the exact same method used in original PER \citep{schaul_quan_antonoglou_silver_2016}. So each sample from the replay buffer would require a total sweep of all transitions to compute the need terms in order to get a sampling distribution. This could be significantly time-consuming, but this sampling scheme still works for our experiment in the Cliffwalk game because the number of states was at most around 10 and the number of experiences in the buffer was at most $10^4$, which still did not require too much time to perform an entire sweep over the replay buffer. More on this computational efficiency problem are discussed in section \ref{discussion}.
\begin{equation*}
    \begin{split}
    P(j) &= \frac{(p_j \cdot Need(s_t, s_j))^\alpha}{\sum_i (p_i \cdot Need(s_t, s_i))^\alpha}\\
    &= \frac{(p_j \cdot M_{tj})^\alpha}{\sum_i (p_i \cdot M_{tj})^\alpha}
\end{split}
\end{equation*}
To verify that the Need term has been learned correctly, we obtained the need values (the expected number of future visits) of every other states given an arbitrary choice of current state 5, as shown in the third graph of Figure \ref{vanilladqn}. We achieved these need values by simply extract the row in the SR matrix $\mathbf{M}$ that corresponds to the fifth state. The graph shows that the need term has been learned properly with a focus on the incoming states, and it was helping the vanilla PER to converge faster. 

Additionally, we also want to investigate the extent to which different implementations could affect the performance of PER. As shown in the last graph of Figure \ref{vanilladqn}, instead of our TD$(\lambda)$ version of the need term, we further investigated two versions that lay on the opposite extremes: the random need and the optimal need. Details of both algorithms are introduced below. 

\textbf{Random Need: }The random need terms (green) was obtained from the definition $M = (I - \gamma T)^{-1}$, where $T$ corresponded to taking random actions (0.25 probability in all four actions) in all states, so that the need terms were totally naive and not at all learned in the training process.

\textbf{Optimal Need: }The optimal need terms (red), however, uses a $T$ that corresponds to the most up-to-date (greedy) policy. In each row of the transition matrix $T_i$, there is only one entry $T_{ij}$ such that $T_{ij} = 1$ and all other entries are all zero because they won't be visited. 

\begin{algorithm*}
	\caption{Prioritized sweeping with Successor Representation (PS-SR)} 
	\begin{algorithmic}[htb]
    \State Initialize $Q(s, a)$, $Model(s, a)$, for all $s$, $a$, and $PQueue$ to empty
    \State Initialize $T$, $\mathbf{M}\leftarrow (I - \gamma T)^{-1}$, $e \leftarrow 0$
    \Loop
         \State $S \leftarrow $ current (nonterminal) state
         \State $A \leftarrow policy(S, Q)$
         \State Take action $A$; observe reward $R$, and state $S'$
         \State $e \leftarrow \gamma\lambda e + \phi(S)$
         \State $\mathbf{M} \leftarrow \mathbf{M} +\alpha e^T\big(\phi(S) + \gamma\phi(S')\mathbf{M} - \phi(S)\mathbf{M}\big)$
         \State $Model(S, A) \leftarrow R, S'$
         \State $P \leftarrow |R + \gamma \max_a Q(S', a) - Q(S, A)|$
         \If{$P > \theta$}
         \State insert $S, A$ into $PQueue$ with priority $P$
         \EndIf
         \State $i = 0$
         \While{$|PQueue| > 0$ and $i < n$}
            \State $i \leftarrow \arg\max_i P_i \cdot \mathbf{M}_{S, S_i}, 1 \leq i \leq |PQueue|$
            \State $S, A \leftarrow pop(PQueue, i)$
            \State $R, S' \leftarrow Model(S, A)$
            \State $Q(S, A) \leftarrow Q(S, A) + \alpha [R + \gamma \max_a Q(S', a) - Q(S, A)]$
            \ForAll{$\bar{S}, \bar{A}$ predicted to lead to $S$}
                \State $\bar{R} \leftarrow $  predicted reward for $\bar{S}, \bar{A}, S$
                \State $P \leftarrow |\bar{R} + \gamma \max_a Q(S, a) - Q(\bar{S}, \bar{A})|$.
                \If{$P > \theta$}
                    \State insert $\bar{S}, \bar{A}$ into $PQueue$ with priority $P$
                \EndIf
            \EndFor
            \State $i \leftarrow i + 1$
        \EndWhile
        \EndLoop
	\end{algorithmic} 
	\label{PSSR}
\end{algorithm*}

\begin{algorithm*}
	\caption{Vanilla Prioritized Experience Replay with Deep Successor Representation (vanilla PER-SR)} 
	\begin{algorithmic}[htb]
		\State \textbf{Input:} minibatch $k$, step-size $\eta$, replay period $K$ and size $N$, exponents $\alpha$ and $\beta$, budget $T$.
		\State Initialize replay memory $\mathcal{H}=\emptyset$, $\Delta = 0$, $p_1 = 1$.
		\State Initialize $T$, $\mathbf{M}\leftarrow (I - \gamma T)^{-1}$, $e \leftarrow 0$.
		\State Observe $S_0$ and choose $A_0 \sim \pi_\theta(S_0)$.
		\For{\textbf{$t=1$ to $T$}}
		    \State Observe $S_t$, $R_t$, $\gamma_t$.
		    \State $e \leftarrow \gamma\lambda e + \phi(S_{t-1})$
            \State $\mathbf{M} \leftarrow \mathbf{M} +\alpha e^T\big(\phi(S_{t-1}) + \gamma\phi(S_t)\mathbf{M} - \phi(S_{t-1})\mathbf{M}\big)$
		    \State Store transition $(S_{t-1}, A_{t-1}, R_t, \gamma_t, S_t)$ in $\mathcal{H}$ with maximal priority $p_t = \max_{i < t}p_i$.
		    \If{$t\equiv 0 \mod K$}
		        \For{\textbf{$j = 1$ to $k$}}
		            \State Sample transition $j \sim P(j) = (p_j \cdot Need(S_{t-1}, S_j))^\alpha / \sum_i (p_i \cdot Need(S_{t-1}, S_i))^\alpha$
		            \State Compute importance-sampling weight $w_j = (N\cdot P(j)) ^ {-\beta} / \max_i w_i$
		            \State Compute TD-error $\delta_j = R_j + \gamma_j Q_{\text{target}}(S_j, \arg \max_a Q(S_j, a)) - Q(S_{j-1}, A_{j - 1})$
		            \State Update transition priority $p_j \leftarrow |\delta_j|$
		            
		            \State Accumulate weight-change $\Delta \leftarrow \Delta + w_j \cdot \delta_j \cdot \nabla_\theta Q(S_{j - 1}, A_{j - 1})$
		        \EndFor
		        \State Update weights $\theta \leftarrow \theta + \eta \cdot \Delta$, reset $\Delta = 0$.
		        \State From time to time copy weights into target network $\theta_{\text{target}}\leftarrow \theta$
		    \EndIf
		    \State Choose action $A_t \sim \pi_\theta(S_t)$
		\EndFor
	\end{algorithmic} 
	\label{vanilla PER algo}
\end{algorithm*}

\begin{algorithm*}
	\caption{Prioritized Experience Replay with Deep Successor Representation (PER-SR)} 
	\begin{algorithmic}[htb]
		\State \textbf{Input:} minibatch $k$, step-size $\eta$, replay period $K$ and size $N$, exponents $\alpha$ and $\beta$, budget $T$.
		\State Initialize replay memory $\mathcal{H}=\emptyset$, $\Delta = 0$, $p_1 = 1$, $\Delta_{\text{sr}} = 0$.
		\State Observe $S_0$ and choose $A_0 \sim \pi_\theta(S_0)$.
		\For{\textbf{$t=1$ to $T$}}
		    \State Observe $S_t$, $R_t$, $\gamma_t$.
		    \State Store transition $(S_{t-1}, A_{t-1}, R_t, \gamma_t, S_t)$ in $\mathcal{H}$ with maximal priority $p_t = \max_{i < t}p_i$
		    \State Calculate SR-vector $m_{t-1} = u_{\theta_{\text{sr}}}(f_{\theta_{\text{sr}}}(S_{t-1}), A_{t-1})$.
		    \If{$t\equiv 0 \mod K$}
		        \For{\textbf{$j = 1$ to $k$}}
		            \State Sample transition $j \sim P(j) = p_j^\alpha / \sum_i p_i^\alpha$
		            \State Compute importance-sampling weight $w_j = (N\cdot P(j)) ^ {-\beta} / \max_i w_i$
		            \State Compute TD-error $\delta_j = R_j + \gamma_j Q_{\text{target}}(S_j, \arg \max_a Q(S_j, a)) - Q(S_{j-1}, A_{j - 1})$
		            \State Update transition priority $p_j \leftarrow |\delta_j|$
		            \State Compute SR-feature $\phi_{j-1} = f_{\theta_{\text{sr}}}(S_{j - 1})$, $\phi_{j} = f_{\theta_{\text{sr}}}(S_{j})$.
		            \State Compute SR-vector $m_{j-1} = u_{\theta_{\text{sr}}}(\phi_{j-1}, A_{j-1})$.
		            \State Compute SR-vector $m_{j} = u_{\theta_{\text{sr}}}(\phi_j, \arg \max_a Q(S_j, a))$
		            \State Compute Need $n_j = (m_{t-1}\cdot\phi_{j-1}) / \left\Vert\phi_{j-1}\right\Vert^2 - \min(0, \min_i m_{i-1}\cdot\phi_{i-1} / \left\Vert\phi_{i-1}\right\Vert^2)$ 
		            \State Get SR loss  $\ell = \left\Vert(s_{j-1} - g_{\theta_{\text{sr}}}(\phi_{j-1})) \right\Vert^2 + \left\Vert \big( \phi_{j-1} + \gamma \cdot m_{j} - m_{j-1} \big)\right\Vert^2$.
		            \State Accumulate weight-change $\Delta \leftarrow \Delta + w_j \cdot \delta_j \cdot n_j \cdot \nabla_\theta Q(S_{j - 1}, A_{j - 1})$
		            \State Accumulate weight-change $\Delta_{\text{sr}} \leftarrow \Delta_{\text{sr}} + \nabla_{\theta_{\text{sr}}}\ell$
		        \EndFor
		        \State Update weights $\theta \leftarrow \theta + \eta \cdot \Delta$, $\theta_{\text{sr}} \leftarrow \theta_{\text{sr}} + \eta \cdot \Delta_{\text{sr}}$, reset $\Delta = 0$, $\Delta_{\text{sr}} = 0$.
		        \State From time to time copy weights into target network $\theta_{\text{target}}\leftarrow \theta$
		    \EndIf
		    \State Choose action $A_t \sim \pi_\theta(S_t)$
		\EndFor
	\end{algorithmic} 
	\label{PERSR algo}
\end{algorithm*}

\begin{figure*}[ht]
\vskip 0.2in
\begin{center}
\centerline{\includegraphics[width=\textwidth]{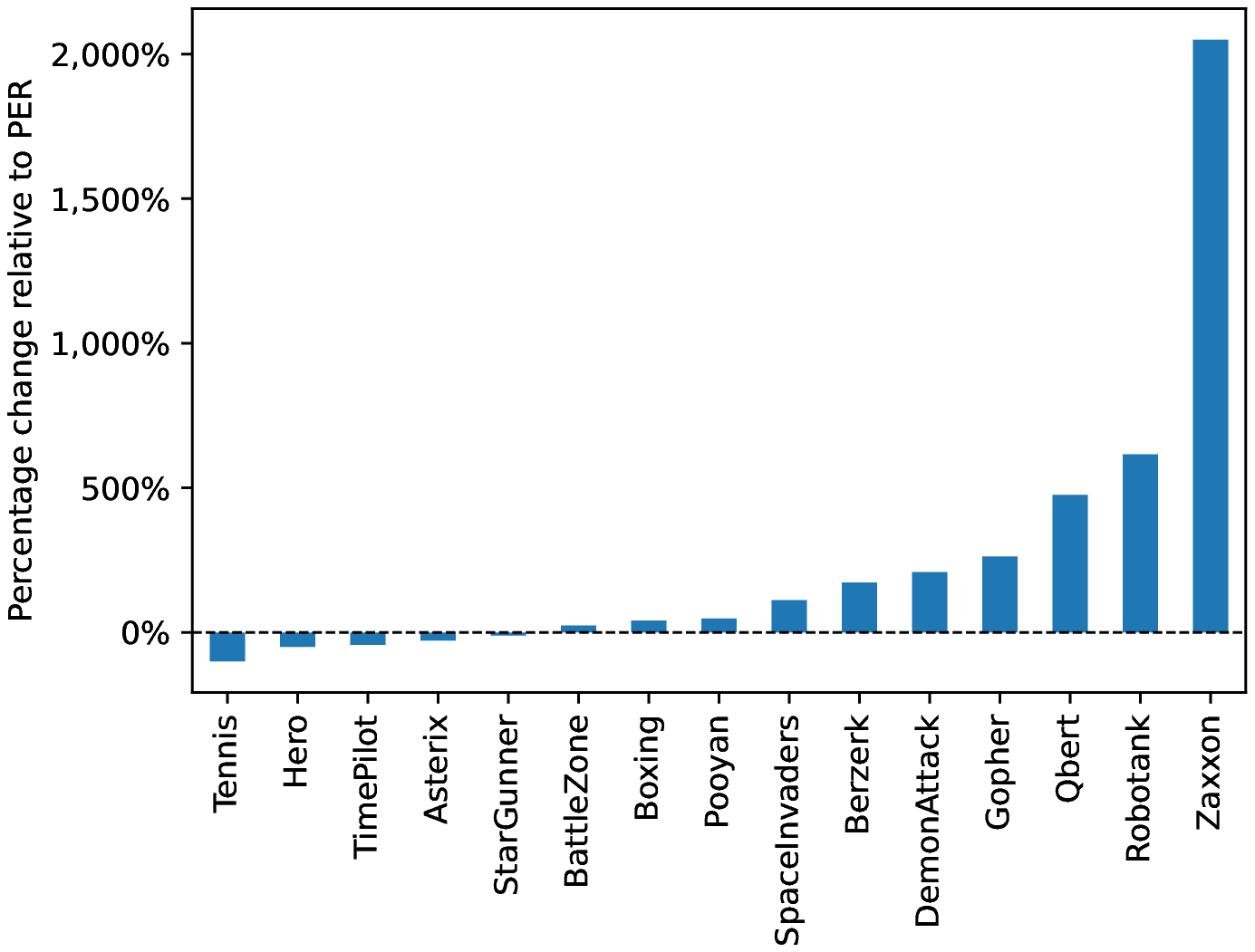}}
\caption{The detailed, enlarged figure of Figure \ref{perfcomp}. The relative performance on a set of selected games of PER-SR with regard to original PER. The games were selected from the 49 games in the original PER paper. The dotted line means the same performance as original PER. Any bar that goes above the line indicates a performance improvement of PER-SR over PER.}
\end{center}
\vskip -0.2in
\end{figure*}

\begin{figure*}[ht]
\vskip 0.2in
\begin{center}
\centerline{\includegraphics[width=1.0\textwidth]{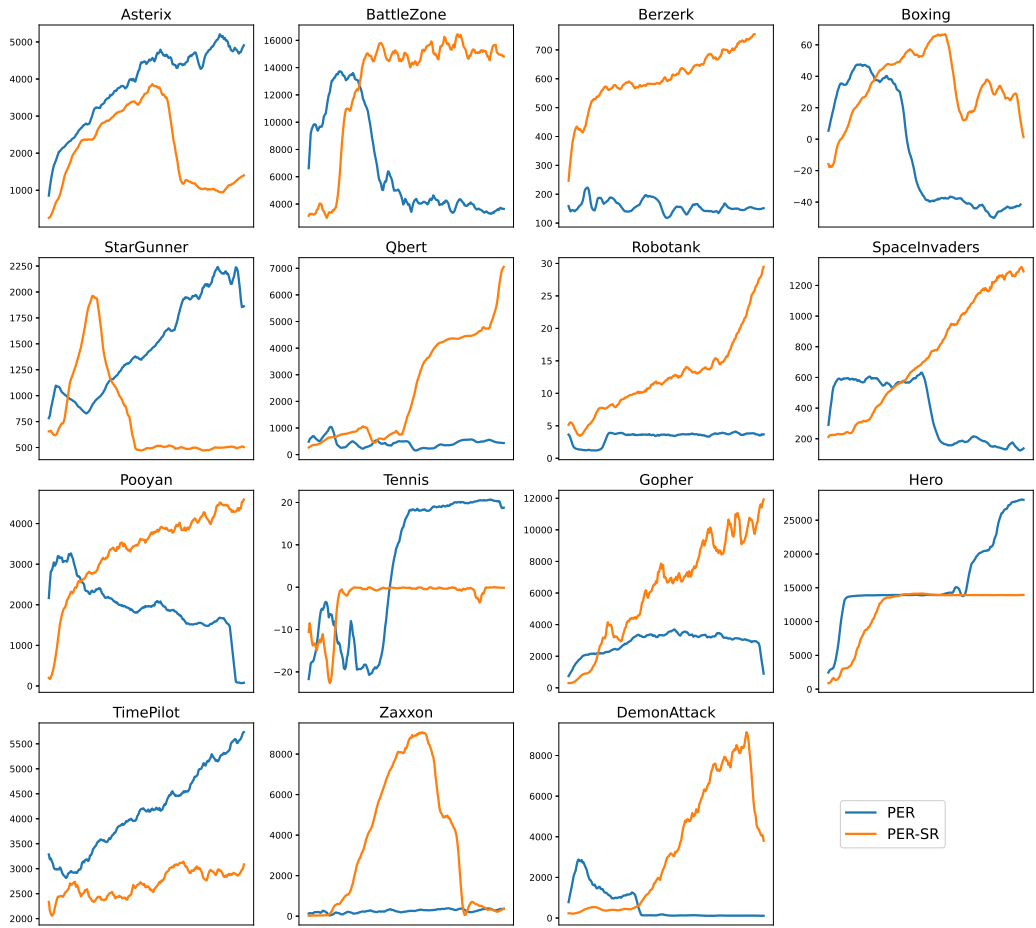}}
\caption{The average game score at each iteration (1 million frames) in a training of 200 million frames in total. The curves were processed in a moving average of 10 million steps for better readability.}
\label{atari}
\end{center}
\vskip -0.2in
\end{figure*}

\end{document}

%% file: math_commands.tex
\usepackage{amsmath,amsfonts,bm}

\def\eqref#1{equation~\ref{#1}}

\def\1{\bm{1}}

\DeclareMathAlphabet{\mathsfit}{\encodingdefault}{\sfdefault}{m}{sl}
\SetMathAlphabet{\mathsfit}{bold}{\encodingdefault}{\sfdefault}{bx}{n}

